\documentclass[10pt,twocolumn,letterpaper]{article}
\usepackage{iccv}

\definecolor{iccvblue}{rgb}{0.21,0.49,0.74}
\usepackage[pagebackref,breaklinks,colorlinks,allcolors=iccvblue]{hyperref}

\usepackage[accsupp]{axessibility}  

\usepackage{amsmath,amsfonts}
\usepackage{array}
\usepackage{amsthm}
\usepackage{textcomp}
\usepackage{stfloats}
\usepackage{url}
\usepackage{verbatim}
\usepackage{graphicx}
\usepackage{adjustbox}
\usepackage{bbm} 
\usepackage{color}
\usepackage{xcolor}
\usepackage{booktabs}
\usepackage{multirow}
\usepackage{enumitem}
\usepackage{colortbl} 
\usepackage[ruled]{algorithm2e}
\SetKwComment{Comment}{/* }{ */}
\usepackage{pifont} 
\definecolor{LightGray}{gray}{0.95}

\usepackage{hyperref}
\hypersetup{
    colorlinks=true,
    linkcolor=blue, 
    urlcolor=blue,  
    citecolor=purple, 
}

\renewcommand{\eqref}[1]{\mbox{Eqn.~\ref{#1}}}

\newcommand{\violet}[1]{\textcolor{magenta}{#1}}

\theoremstyle{remark}
\newtheorem{remark}{Remark}

\begin{document}

\title{ELIP: Efficient Discriminative Language-Image Pre-training with Fewer Vision Tokens}

\author{%
  Yangyang $\text{Guo}^1$, Haoyu $\text{Zhang}^1$, Yongkang $\text{Wong}^1$, Liqiang $\text{Nie}^2$, Mohan $\text{Kankanhalli}^1$ \\
  $^1$National University of Singapore, $^2$Harbin Institute of Technology (Shenzhen) \\
}



\maketitle

\begin{abstract}
Learning a versatile language-image model is computationally prohibitive under a limited computing budget.
This paper delves into \emph{efficient discriminative language-image pre-training}, an area that has received relatively little attention despite its importance in reducing computational cost and carbon footprint.
To that end, we propose a vision token pruning and merging method ELIP, to remove less influential tokens based on the supervision of language.
Our method is designed with several advantages, such as being computation-efficient, memory-efficient, and trainable-parameter-free, and is distinguished from previous vision-only token pruning approaches by its alignment with pre-training task objectives.
We implement this method in a progressive merging manner using several sequential blocks.
To evaluate its generalization performance, we apply ELIP to five commonly used discriminative language-image pre-training models.
The models considered include contrastive-only pre-training methods (\eg CLIP) and enhanced cross-modal learning frameworks (\eg BLIP).
We then utilize public image-caption datasets with 4M to 12M images for pre-training.
Our experiments demonstrate that even with the removal of $\sim$30$\%$ vision tokens across twelve ViT layers, ELIP maintains comparable performance with baselines ($\sim$99.2\% on average) over various downstream tasks, including image classification (zero-shot, linear probing, and full fine-tuning), cross-modal retrieval, VQA, image captioning, \emph{etc}.
In addition, the spared GPU resources by our ELIP allow us to scale up to larger batch sizes given a fixed budget, thereby accelerating model pre-training and even sometimes enhancing downstream model performance.
Our code has been \href{https://github.com/guoyang9/ELIP}{open sourced}.
\end{abstract}


\section{Introduction}
Recent advancements over various benchmarks are primarily due to large model pre-training~\citep{qwen3, chameleon}.
These pre-trained models stand out for their versatility and generalization ability, and are further aided by the scaling law~\citep{scaling_law1,scaling-law2},
which states that expanding model size, computing budget, and training data leads to increasingly better performance.
Nevertheless, the use of pre-trained large models often incurs a noticeable carbon footprint and faces great challenges for deployment in resource-constrained environments. 
As a result, many efforts have been devoted to optimizing the efficiency-effectiveness trade-off of large models~\citep{adapter, prompt, distill-bert}.

\begin{figure}[t!]
  \centering
  \includegraphics[width=1.0\linewidth]{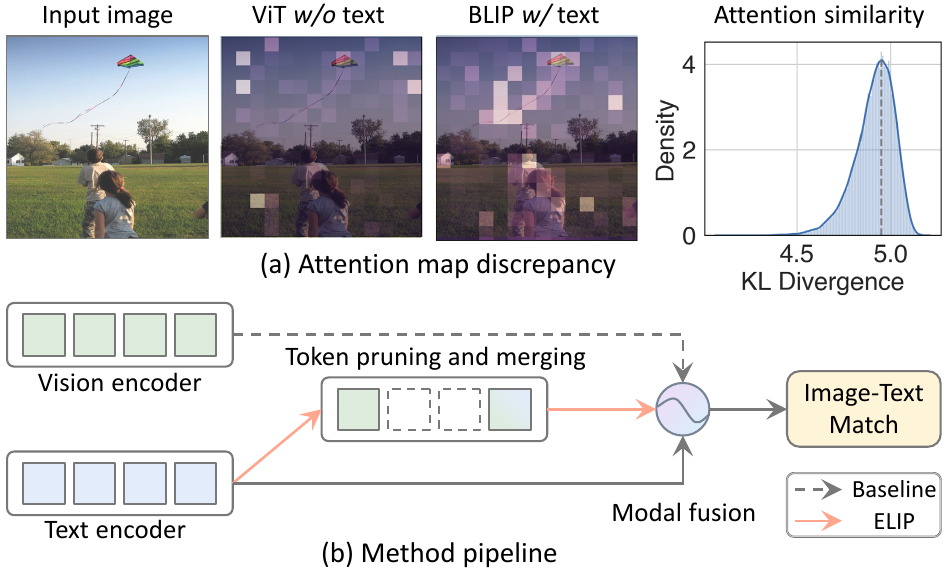}
  \caption{Visualization of attention map discrepancy between vision-only ViT and vision-language BLIP models and pipeline of our proposed method ELIP.
  (a) When presented with the same image, ViT and BLIP often see different regions, resulting in a large KL divergence of their attention maps.
  (b) ELIP achieves efficient discriminative language-image pre-training by pruning less important vision tokens supervised by the text.
  }\label{fig:teaser}
\end{figure}

Conventional efficient learning approaches, \eg knowledge distillation~\citep{pkd, kd}, low-rank approximation~\citep{low-rank, pela}, and quantization~\citep{quan3, quantization-new}, are commonly employed to compress a cumbersome model into a lightweight one.
By this means, the computational overhead and memory cost are reduced, despite the complexity involved in developing these compression algorithms.
Since the emergence of Vision Transformers (ViTs)~\citep{vit}, much recent focus has been tailored to a more explainable and effective approach, \ie \emph{vision token pruning}. 
ViTs embed images using non-overlapped patches, which is distinct from the traditional approach of CNNs that explicitly incorporates spatial inductive bias~\citep{vit-proof, convnext}.
This operation often leads to redundant vision tokens that can be safely removed without significantly compromising models' accuracy~\citep{vit-prune-class, vit-prune3, vit-prune4}.
However, existing pruning methods in the vision-only domain \textbf{predominantly} rely on an objective-free approach (\eg ToMe~\citep{merge} and ~\citep{vit-efficient-pami}), whereby the pruning mask is learned from signals of current or preceding layers~\citep{vit-prune-class, vit-prune4}. 
This approach may entail the risk of removing tokens that play a crucial role in achieving the task objective, especially for vision-language models (see Fig.~\ref{fig:teaser} for an example).

We notice that there is relatively sparse work on efficient discriminative language-image pre-training\footnote{Unlike generative models, discriminative language-image models do not utilize large language models (LLMs) as their final interface.}.
In general, the natural correspondence between language and image mutually describes which token(s) are dispensable for training a generalizable multi-modal model.
Besides, recent methods often employ separate pre-trained encoders for the two input modalities, wherein the encoding operation is asynchronous\footnote{The complex parallel computing, though feasible, usually prohibits researchers from encoding language and image simultaneously.}.
This allows us to leverage the output from the text encoder as supervision for removing vision tokens (refer to Fig.~\ref{fig:teaser}(b)), which differs significantly from the vision-only domain (In addition, the vision-only model and language-image model usually concentrate on different regions, as shown in Fig.~\ref{fig:teaser}(a)).
The language tokens, on the other hand, are less redundant in their representation due to short context (\eg 20 words per sentence) and high information density~\citep{maskAE}.
We therefore only ablate language token pruning for completeness~\citep{flip}.

Our method does not require \emph{any incremental trainable parameters} beyond backbone language-image models.
Building on the observation that the attention map on vision tokens becomes increasingly concentrated with deeper layers (see Fig.~\ref{fig:depth}), we implement vision token pruning and merging in a progressive and multi-stage way.
We integrate our ELIP as a plug-and-play module into \emph{five} popular discriminative language-image pre-training models.
More specifically, CLIP ViT-S/16 and CLIP ViT-B/16~\citep{clip} are representative of contrastive-only models, whereas ALBEF~\citep{albef}, BLIP~\citep{blip}, and METER~\citep{meter} fall under the category of enhanced cross-modal learning models.
We pre-train them from scratch on datasets with image-caption pairs using 4M to 12M images, where the datasets consist of MSCOCO Caption~\citep{coco}, Visual Genome~\citep{genome}, SBU~\citep{sbu}, Conceptual Captions~\citep{cc}, and Conceptual-12M~\citep{cc12m}.
Through our experimental results, we demonstrate that removing $\sim$30\% vision tokens can well maintain the model performance (99.2\% on average) on downstream tasks, including image classification (zero-shot, linear probing, and full fine-tuning), image-text retrieval, visual question answering, visual entailment, image captioning, and natural language for visual reasoning.
In addition, the spared GPU memory by our method enables model scaling up with larger batch sizes given a limited budget, and even sometimes slightly boosts downstream model fine-tuning.
We also validate the effectiveness of combining our pre-training method with several parameter-efficient transfer learning approaches.
We believe that our approach provides valuable insights for future discriminative language-image pre-training to develop more advanced models, whilst with significantly reduced computational cost and carbon footprint. 

It is worth noting that our method is orthogonal to recent generative language-image pre-training models such as LLaVA~\citep{llava}, Qwen-VL~\citep{qwen-vl}, and DeepSeek-VL~\citep{deepseek-vl2}.
While these models demonstrate strong performance, we argue that discriminative language-image pretraining remains advantageous in two key aspects:
1) superior parameter-efficiency and 2) greater reliability for discriminative tasks, as generative models often require additional response parsing, which can lead to unstable performance~\citep{discriminative-cvpr};
More importantly, discriminative pretraining forms the foundation for many generative models; for instance, the vision encoder from CLIP is widely adopted by recent Vision LLMs~\citep{llava}.
Therefore, advancing efficient discriminative language-image pretraining is both significant and essential.

In summary, the contribution of this work is three-fold:
\begin{itemize}[leftmargin=*]
    \item This paper presents a comprehensive study on vision token pruning guided by text supervision in discriminative language-image pertaining.
    Our proposed method can be seamlessly applied to both contrastive-only and enhanced cross-modal learning models.
    \item We offer a detailed explanation of why the proposed method is effective, supported by both in-depth analysis and empirical results.
    \item We conduct extensive downstream evaluations across a diverse set of tasks, covering a total of \emph{eleven} datasets.
    The results show that our method introduces minimal performance degradation while achieving memory and time efficiency when pruning approximately 30\% of vision tokens.
\end{itemize}

\section{Related Work}\label{sec:related_work}
\subsection{Discriminative Vision-Language Transformers}
The past few years have witnessed the popularity of Transformers in natural language processing and computer vision~\citep{transformer, vit}.
Given its overwhelming performance in these related domains, researchers have actively extended this technique to vision-language tasks.
In particular, a \emph{pre-train then fine-tune} paradigm is adopted by mainstream methods and the models are often pre-trained on certain large-scale vision-language datasets~\citep{bert, llava, qwen-vl}.

Unlike previous single-modality model pre-training, the vision-language domain requires two heterogeneous inputs.
The ubiquitous image-text pairs, \ie textual caption of an image, serve as the key data format for pre-training due to their easy availability. 
Common datasets include Conceptual Captions~\citep{cc}, Visual Genome~\citep{vg}, COCO Captions~\citep{coco}, and LAION-400M~\citep{laion}.
At the bedrock of vision-language Transformers lies the embedding behavior of the two modalities. 
Pertaining to the vision embedding, the feature extraction has grown from grid~\citep{VQA1}, region features~\citep{bottom-vqa} of CNN models, to the recent patch features of Transformers~\citep{vl-survey}.
In contrast, the text tokenization promptly changed from traditional Word2Vec to BERT-style pre-trained embeddings after the prevalence of modern language modeling~\citep{transformer, bert}.
On top of the embedding process, there are generally two types of pre-training approaches: contrastive-only and enhanced cross-modal learning.
The former, exemplified by CLIP~\citep{clip}, employs a late fusion strategy in which visual and textual inputs are encoded independently before being combined through a fusion operation~\citep{lxmert, villa}, typically optimized using a contrastive pretraining objective.
Methods from the latter enhance the contrastive pre-training with cross-modal goals, such as masked language modeling~\citep{bert, vil-bert}, masked vision modeling~\citep{lxmert, maskAE}, and image-text matching~\citep{blip, blip2}.

\subsection{Network Pruning}
Network pruning is used to remove unnecessary or less important components in models~\citep{prune-new-1, prune-new-2, prune-new-3}. 
By removing some connections or parameters, the original dense network reduces to a sparse one, in which the required capacity for storage will dwindle as well as the volume of computations~\citep{prune-structured, prune-unstructured}.
Based on the granularity of reduction, existing methods can be roughly grouped into unstructured pruning and structured pruning.
The former refers to pruning less salient components, such as neurons or connections between layers~\citep{prune-unstructured, prune-unstructured3}.
In contrast, the latter aims to remove a large bundle of parameters~\citep{prune-structured, prune-new-2}, which we mainly discuss in this section.
Previous structured pruning methods mostly target removing less influential Transformer heads~\citep{prune-head}, layers~\citep{prune-layer}, and convolutional channels~\citep{prune-channel}. 
With the startling success of ViT~\citep{vit}, increasing research has been devoted to pruning input tokens of each layer due to the following two reasons.
First, the input tokens from different layers have different redundancies, and only a small number of them contribute significantly to the accuracy of models~\citep{prune-structured, token-sparse}.
Second, pruning tokens leads to more visual explainability as compared to other elements, such as attention heads or layers.


\subsection{Efficient Learning}
Efficiency has long been an engaging problem across numerous research fields~\citep{pre-dl1, pre-dl2, peft-pami1, peft-pami-2}.
In the deep representation learning era, the progressive improvements often trade increased model complexity, latency, and footprints~\citep{efficient-survey, quantization-new}.
To address these challenges, researchers have primarily focused on three categories of approaches: knowledge distillation (KD), quantization, and low-rank approximation.

Deemed as a principled model compression algorithm, early KD aims to transfer the knowledge from a cumbersome teacher model to a lightweight student model via class logit alignment~\citep{kd, kd-mm}.
Recent focus has been shifted to feature-based knowledge transfer due to its performance advantage over conventional logit-based ones~\citep{fitnet, kd-new, tiny-bert}.
For example,~\citep{pkd, tiny-bert} distilling the knowledge from hidden states and attention matrices, which on the other hand, can further help bypass the logit-free training objectives.
However, choosing which features to align remains challenging as there is no explicit teacher-student layer match from a theoretical basis.
Quantization, from another angle of efficient learning, maps larger bit parameters to smaller ones, \eg 32-bit floating point to an 8-bit integer~\citep{quantization-survey}.
This kind of method is not dependent on model structures, which makes it flexible in various neural networks~\citep{quan3, quantization-new}.
Different from the above two categories, low-rank approximation methods decompose the original matrix into two matrices with a lower rank~\citep{al-bert, pela, low-rank}.
Nevertheless, determining the optimal rank of matrices to achieve the performance-efficiency trade-off has been proven to be rather difficult in the literature~\citep{low-rank2}.

Transformer-based approaches have thrived in diverse domains since their introduction~\citep{transformer, bert}.
These models often involve billions of parameters, which consequently motivates some specific methods working on addressing the parameter efficiency problem~\citep{al-bert}.
The typical strategy is to add a few learnable parameters while freezing the majority of the Transformer backbone during downstream training~\citep{lora, lora-new, adapter-new}.

\section{Method}\label{sec:method}
\subsection{Preliminary}

\begin{figure*}[t!]
  \centering
  \includegraphics[width=1.0\linewidth]{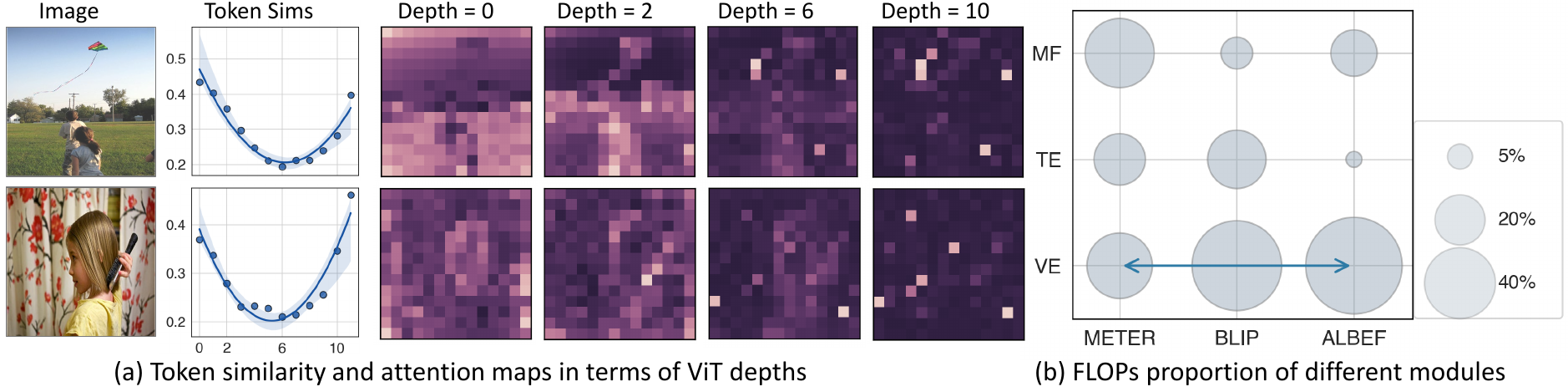}
  \caption{Token similarity and attention maps across different ViT layers of BLIP~\citep{blip}, as well as the FLOPs proportion of different modules for three typical language-image pre-trained models.
  (a) The attention distribution over image tokens grows from uniform to concentrated with layers going deeper.
  Besides, the token similarity initially decreases but then significantly increases, indicating that more vision tokens become redundant.
  (b) Notably, the vision encoder (VE) accounts for the majority of the computational cost of language-image models (compared to the text encoder - TE and modal fusion - MF).
  }\label{fig:depth}
\end{figure*}

\subsubsection{Discriminative Language-Image Pre-Trained Models} \label{sec:preliminary}
Transformers have grown into a fundamental building block of many modern language-image models~\citep{albef, vilt, blip, meter}. 
According to the common training paradigm, existing models can be split into three modules: vision encoder, text encoder, and multi-modal fusion.
In this way, the language and vision encoders can be aligned with separate pre-trained models to facilitate knowledge transfer.

\noindent\textbf{Vision Encoder}.
Recent discriminative language-image pre-training models often employ the ViT model~\citep{vit} as the vision encoder.
ViT first partitions an RGB image $Img \in \mathbb{R}^{3 \times H \times W}$ into $M \times M$ non-overlapping patches.
Together with a class token [CLS], these image patches are thereafter fed into $N$ layers with self-attention as the basic operation, $\mathbf{X}^v \in \mathbb{R}^{(1 + M^2) \times d}$, where $d$ denotes the embedding size.

\noindent\textbf{Text Encoder}.
After tokenizing the input sentence~\citep{bert}, current methods often employ a special [CLS] token at the beginning and a [SEP] token at the end of the sequence. 
These tokens serve to delimit the sentence and enable the text encoder to extract the token encoding, which is represented by $\mathbf{X}^t \in \mathbb{R}^{(2 + T) \times d}$, where $T$ denotes the sentence length.

\noindent\textbf{Modal Fusion}.
The model fusion module employs frameworks that are similar to those utilized by the Transformer decoder~\citep{transformer}. 
In particular, the common practice includes the cross-attention between the vision and text encoders~\citep{albef, blip}, as well as merged attention techniques~\citep{meter}.
Notably, CLIP models~\citep{clip, openclip} simply leverage cosine similarity as the modal fusion module.

\noindent\textbf{Pre-Training Objectives}.
Pre-training of discriminative language-image models on large-scale image-caption datasets~\citep{coco, cc} is made possible by several pretext objectives. 
The most typical one is image-text matching (ITM), which is implemented in a contrastive learning framework. 
For example, CLIP~\citep{clip} utilizes ITM objective only for pre-training. 
The other objectives such as masked language modeling (MLM), aim to reconstruct masked text tokens given the remaining ones and the image feature. 

\subsubsection{Motivation}
While achieving promising results on downstream tasks, discriminative language-image pre-training models can suffer from computational inefficiency. 
To approach this problem, Fig.~\ref{fig:depth} illustrates two critical observations that motivate this work:

\begin{remark}
Fig.~\ref{fig:depth}(b) indicates that the vision encoder usually accounts for the majority of overhead in a language-image model, especially for ALBEF and BLIP.
Given this observation, reducing the computational cost of the vision encoder will yield great improvement in model training efficiency.
\end{remark}

\begin{remark} \label{rmk: depth}
The vision tokens from ViT are redundant in their representations~\citep{vit-prune-class, merge}, as is the case for these language-image models.
Moreover, the attention value distribution becomes increasingly concentrated for deeper ViT layers (as seen in Fig.~\ref{fig:depth}(a)).
One insight from this observation is that we can progressively remove these tokens that are less useful for the image-text matching objective, and therefore obtain better computational efficiency.
\end{remark}

\subsection{Method Architecture} \label{sec:architecture}
In light of the above two observations, this paper aims to study the \emph{efficient discriminative language-image pre-training} by means of \emph{vision token pruning and merging}.
We do not remove image patches in the input space~\citep{flip} as we believe some background patches still provide useful features for cross-modal semantic understanding.
Instead, we propose to prune the vision tokens that are less influential for the matching goal of the given image-text pair.

As noted in Remark~\ref{rmk: depth}, the redundancy of vision tokens increases as the depth of Transformer layers grows.
In view of this, we design a progressive pruning strategy with multiple-stage blocks that follow the hierarchical ViT structures, such as Swin Transformer~\citep{swin} and MetaFormer~\citep{poolformer}.
Specifically, our approach involves dividing a standard ViT encoder into four distinct and non-overlapping blocks:

\begin{itemize}[leftmargin=*]
    \item \textbf{Block I} remains unaltered for the first two ViT layers.
    Unlike the vision-only domain, both causal and background features contribute a lot to the semantic understanding in a language-image model.
    It is thus preferable to leave these layers close to the input unchanged.
    \item \textbf{Block II} consists of two layers and prunes a few of all the vision tokens (\eg 10\%) that are redundant.
    \item \textbf{Block III} removes much more tokens (such as 25\%) preceding the next six layers.
    Fig.~\ref{fig:depth} shows that the attention maps tend to exhibit increasingly concentrated for deeper layers, indicating that the model focuses primarily on a small number of representative image regions.
    \item \textbf{Block IV} further performs token pruning and keeps $\alpha$ (\eg 40$\%$) of all vision tokens with the last two layers, which are crucial for the subsequent multi-modal fusion.
    Inspired by MaskAE~\citep{maskAE}, we demonstrate that we can maintain a comparable fine-tuning model performance by retaining only a small group of vision tokens.
\end{itemize}
Based on this architecture, in the next subsection, we will elaborate more on the pruning and merging details of less important vision tokens in each block.

\subsection{Vision Token Pruning and Merging}
Fig.~\ref{fig:teaser} illustrates the encoding process typically used in discriminative language-image models~\citep{albef, meter}, in which text and image inputs are processed separately. 
This non-parallel operation allows us to use the output of the text encoder to help remove irrelevant tokens in the vision encoder, which can provide significant benefits over vision-only pruning models~\citep{merge, vit-prune-class}. 
Moreover, the alignment between the image and text is determined by the features extracted from the [CLS] token.
As a result, we employ the fusion of these two sets of features to jointly decide which tokens are influential for each given block.
We outline the process of the algorithm in Alg.~\ref{alg:prune}.
Specifically, the number of tokens reduces from 1 + $M_i$ to 2 + $\alpha M$ for block $i$ with the pruning and merging approach.
Here, $M_i$ and $M$ represent the token numbers of the current block and the foremost input, respectively.
The retaining ratio $\alpha$ is defined in Sec.~\ref{sec:architecture} and is always less than 1.0, \eg 0.4 for the last block.


\begin{algorithm}
\caption{Vision token pruning and merging for ELIP} \label{alg:prune}

\KwData{Vision token feature from the $i$-th block $\mathbf{X}^{v, i} \in \mathbb{R}^{(1 + M_i) \times d}$,   
text feature $\mathbf{X}^t_{[CLS]} \in \mathbb{R}^{d}$ from the [CLS] token, 
feature combination coefficient $\lambda$, 
number of the overall vision tokens $M$ and the retaining ratio $\alpha$}

\KwResult{Pruned vision token feature $\mathbf{X}^{v, i+1} \in \mathbb{R}^{(2 + \alpha M) \times d}$}
$\mathbf{X}^{v, i}_{[CLS]} \gets \lambda \mathbf{X}^{v, i}_{[CLS]} + (1-\lambda)\mathbf{X}^t_{[CLS]}$\; 
\For{each layer in the current block}{
forward $\mathbf{X}^{v, i}$ until the end of this layer\;
\If{is\_last\_layer}{
save the attention value $\mathbf{\xi} \in \mathbb{R}^{M_i}$ for pruning \Comment*[r]{pooled attention of the [CLS] token}
}
}
$\mathbf{\xi}$.sort(\textcolor{blue}{descending=True}) \Comment*[r]{excluding the [CLS] token}
$\overline{\mathbf{X}}^{v, i} \in \mathbb{R}^{(1 + \alpha M) \times d} \gets \text{tokens with top $\alpha M$ attention values}$\;
\If{is\_merge}{ 
$\hat{\mathcal{V}_i} \gets \mathcal{V}_i \setminus \{\text{top-}\alpha M (\mathbf{\xi})\} $\;
$\hat{\mathbf{\xi}} \gets \text{normalizing remaining $M_i - \alpha M$ values}$\;
$\mathbf{X}^{v, i}_{merge} \gets \sum_{k \in \hat{\mathcal{V}}_i} \mathbf{X}^{v, i}_k \hat{\mathbf{\xi}}_k$  \Comment*[r]{weighted sum of the remaining tokens}
}
$\mathbf{X}^{v, i+1} \in \mathbb{R}^{(2 + \alpha M) \times d} \gets concat(\overline{\mathbf{X}}^{v, i}; \mathbf{X}^{v, i}_{merge})$
\end{algorithm}

To this end, we first replace the image [CLS] token features with the fusion of itself and the text [CLS] features,
\begin{equation} \label{eqn:lambda}
    \mathbf{X}^{v, i}_{[CLS]} = \lambda \mathbf{X}^{v, i}_{[CLS]} + (1-\lambda)\mathbf{X}^t_{[CLS]},
\end{equation}
where $\lambda$ is a coefficient hyperparameter balancing the contribution of vision and text token features.
In the next step, we perform token pruning and merging without considering gradients, \ie in a \emph{stop-gradient} fashion.
All the vision token features $\mathbf{X}^{v, i}$ are thereafter fed to each layer of the current block and only the attention values of [CLS] from the final layer, namely $\mathbf{\xi}$ are preserved,
\begin{equation}
    \mathbf{X}^{v, i}, \mathbf{\xi} = \text{BLOCK}_i (\mathbf{X}^{v, i}; \mathbf{\Theta}),
\end{equation}
where $\Theta$ represents the involved parameters and with no gradient during this computation.
We then retain those token features based on a pre-defined retaining ratio $\alpha$. 
\begin{equation}
    \overline{\mathbf{X}}^{v, i} = concat(\mathbf{X}^{v, i}_{[CLS]}; \{\mathbf{X}^{v, i}_{j}\}_{j \in \{\text{top-}\alpha M (\mathbf{\xi})\}}), 
\end{equation}
where $\text{top-}n()$ denotes the index set with the largest $n$ values.  
For the remaining tokens, we merge them into a single token according to their attention after re-normalization,
\begin{equation} \label{eqn:merge}
    \begin{cases}
    \hat{\mathcal{V}}_i = \mathcal{V}_i \setminus \{\text{top-}\alpha M (\mathbf{\xi})\}, \\
    \hat{\mathbf{\xi}} = norm(\{\mathbf{\xi}_{j}\}_{j \in \hat{\mathcal{V}}_i}),\\
    \mathbf{X}^{v, i}_{merge} = \sum_{k \in \hat{\mathcal{V}}_i} \mathbf{X}^{v, i}_k \hat{\mathbf{\xi}}_k,
\end{cases} 
\end{equation}
where $\mathcal{V}_i$ represents the overall token index set from $1\rightarrow M_i$ of the current block $i$.
We finally concatenate it with the remaining tokens after pruning.
This approach ensures that the subsequent ViT layers will consider a smaller number of tokens, leading to more efficient processing.

\subsection{Method Analysis}

\subsubsection{In-Depth Understanding}
In this section, we provide a formal formulation of vision token pruning for discriminative language-image pre-training. 
Specifically, we first formalize the objective of vision token pruning as:
\begin{equation}
    \underset{\mathcal{V}_s \subseteq \mathcal{V}_f, |\mathcal{V}_s| \leq \alpha |\mathcal{V}_f|}{\max} \text{Perf}(\mathcal{V}_s),
\end{equation}
where $\mathcal{V}_f$ and $\mathcal{V}_s$ represent the full vision token set and selected token set, respectively; $\text{Perf}$ refers to the downstream model performance, and $\alpha$ is the token retaining ratio.
Since it is infeasible to track all downstream tasks in real time, we instead optimize the following objective at the pre-training stage: 
\begin{equation}
    \underset{\mathcal{V}_s \subseteq \mathcal{V}_f, |\mathcal{V}_s| \leq \alpha |\mathcal{V}_f|}{\min} \mathcal{L} (f(\mathcal{V}_s; \Theta), y),
\end{equation}
where $\mathcal{L}$ represents the pre-training loss function and $y$ refers to the pre-training target as described in Sec.~\ref{sec:preliminary}. We relate this optimization problem to the lottery ticket hypothesis~\citep{lottery} from a token pruning perspective. In specific, we posit that there exists a subset $\mathcal{V}_s \subseteq \mathcal{V}_f$ such that
\begin{equation}
    \mathcal{L} (f(\mathcal{V}_s; \Theta), y) \leq \mathcal{L} (f(\mathcal{V}_f; \Theta), y) + \epsilon,
\end{equation}
for some small acceptable degradation $\epsilon$, and our vision token pruning essentially searches for this subset. Our results in Sec.~\ref{sec:experiment} further show that we can indeed find such $\mathcal{V}_s$ with minor perturbations or degradations in pre-training losses, while still maintaining comparable performance across diverse downstream tasks.

\noindent\textbf{Why text supervision}? The contrastive language image pre-training can be generally conceptualized as maximizing the mutual information between image $V$ and text $T$:
\begin{equation}
    \underset{\Theta}{\max} I (V; T),
\end{equation}
where $I (V; T)$ denotes the mutual information between the given image and text pair. 
In theory, mutual information measures the dependency between two variables, such as image and text representations in our case, and is defined as:
\begin{align*}
    \underset{\Theta}{\max}I (V; T), \\
    \rightarrow \underset{\Theta}{\max} D_{KL} (p(V, T) || p(V)p(T)), \\
    \rightarrow \underset{\Theta}{\max} D_{KL} (\textcolor{blue}{p(V|T)} p(T) || p(V)p(T)). 
\end{align*}
Previous vision-only pruning methods typically ignore the conditional distribution $\textcolor{blue}{p(V|T)}$, which captures the relevance of visual tokens given the accompanying text. 
In contrast, we find that incorporating this information during pruning leads to increased mutual information between visual and textual modalities, thereby enhancing the effectiveness of discriminative language-image pretraining.
    
\subsubsection{Relation to Other Methods}
Our method utilizes the multi-modal information (\ie weighted sum) for pruning, as shown in Eqn.~\ref{eqn:lambda}.
We illustrate two extreme cases where $\lambda$ takes the values of 0 or 1.
On the one hand, when $\lambda=0$, our method degrades to vision-only pruning, wherein there is no supervision from the text.
On the other hand, when $\lambda=1$, similar to TRIPS~\citep{trips}, the pruning is solely determined by the text, resulting in a significant drop in model performance.
We relate this result to that of the momentum update in MoCo~\citep{moco}.
Specifically, replacing the vision [CLS] with the text [CLS] introduces a substantial modality gap, which confuses the model's learning process in terms of which tokens it should focus on.
In contrast, a slowly evolving vision [CLS] ($0<\lambda<1$) serves as the core to gracefully maintain modality consistency.

\subsubsection{Complexity Analysis}
To explain the efficiency of our method, let us consider the ViT-Base model, which contains a total of $\pi$ vision tokens. 
Given that the ViT model comprises twelve layers, the overall memory complexity can be roughly estimated as $\mathcal{O}(f(12 \times \pi))$, where $f$ denotes the token processing in a single layer.
Let us assume the retaining ratios are 90\%, 65\%, and 40\% for the last three blocks\footnote{The ratios are defined based on the total number of tokens.}, respectively.
With our ELIP method, the resulting memory complexity with respect to tokens is reduced to $\mathcal{O}(f((2 + 2 \times 0.9 + 6 \times 0.65 + 2 \times 0.4) \times \pi)) \approx \mathcal{O}(f(8.5 \times \pi))$, which corresponds to a reduction of approximately 30\% in memory usage relative to the original baseline model. 
This reduction enables the pre-training of language-image models using larger batch sizes or deeper layers, while also reducing the computational complexity as fewer tokens are taken in the self-attention operation.

\section{Experiments}\label{sec:experiment}
We applied our method to two categories of discriminative language-image pretraining models: contrastive-only and enhanced cross-modal learning.
We prioritize experiments on the latter group for two reasons: (1) the training objectives of enhanced cross-modal models subsume those of contrastive-only models, and (2) they support a wider range of downstream tasks.

\subsection{Common Experimental Settings}
\subsubsection{Compared Baselines}
Since efficient discriminative language-image models are quite sparse in the literature, we adapted two SOTA vision token pruning baselines in the vision-only domains for comparison: \textbf{EViT}~\citep{vit-prune-class}, METER~\citep{meter}, and \textbf{ToMe}~\citep{merge}.
Both methods prune the vision tokens for each ViT layer in an unsupervised manner.

\noindent \textbf{EViT}~\citep{vit-prune-class}
involves reorganizing the vision tokens in ViT based on the learned importance score from the [CLS] token. 
We followed the original source code to progressively prune and fuse the inattentive vision tokens. 
Specifically, we employed the default configuration with a keep rate of 0.7 and excluded the gradual keep rate decaying due to the training epoch inconsistency (\ie 15 of ELIP \textit{v.s.} 300+ of EViT). 
We integrated this baseline method into the Transformer block for CLIP ViT-S/16, CLIP ViT-B/16, ALBEF and BLIP and wrapped it within the ResidualAttentionBlock module for METER.

\noindent \textbf{ToMe}~\citep{merge}
first identifies similar tokens and then merges them to reduce the number of vision tokens. 
We used their official public code to implement the proportional attention mechanism in the self-attention module of ViT.
In our experiments, we set the number of tokens reduced per layer $r$ to the default value of 13. 
Additionally, following the practice of ToMe, we set ``prop\_attn'' to be true to ensure that merged tokens can receive proportional attention.
\emph{We applied ToMe to ALBEF and BLIP while excluding CLIP ViT-S/16, CLIP ViT-B/16, and METER due to the distinct model architecture gaps}.

\subsubsection{Downstream Task Configurations}
We strictly followed the experimental settings used by all five discriminative language-image pre-training models and kept most of them untouched.
We employed a smaller batch size and reproduced the results of the baseline models.
For more information on the exact batch size configurations that were employed, please refer to our released code.

\begin{table*}[htbp]
    \centering
    \caption{Comparison of text retrieval (TR) and image Retrieval (IR) performance on Flickr30K and MSCOCO datasets.
    Metrics pertaining to efficiency are estimated for pre-training rather than downstream fine-tuning.
    The TFLOPs calculation is based on a batch size of 36, and the memory usage estimates are only applicable to the tested backbones and our proposed methods.
    RRT: Ratio of Removed Tokens; LTY: Latency (minutes) per epoch; Mem: Memory (GB).} 
    \begin{adjustbox}{width=\textwidth}
    \begin{tabular}{l|cccc *{4}{|ccc}}
    \toprule
        \multirow{3}{*}{Model}  & \multicolumn{4}{c|}{\violet{Effiency-Related}}     & \multicolumn{6}{c|}{Flickr30K}                        & \multicolumn{6}{c}{MSCOCO}                        \\
                                \cmidrule(lr){2-5}                          \cmidrule(lr){6-11}                                     \cmidrule(lr){12-17}
                                & \multirow{2}{*}{\violet{RRT$\uparrow$} } & \multirow{2}{*}{\violet{TFLOPs$\downarrow$}}   & \multirow{2}{*}{\violet{LTY$\downarrow$}}  
                                                                                                    & \multirow{2}{*}{\violet{Mem$\downarrow$}}        
                                                                            & \multicolumn{3}{c|}{TR}   &\multicolumn{3}{c|}{IR}    & \multicolumn{3}{c|}{TR}   & \multicolumn{3}{c}{IR} \\
                                                                            \cmidrule(lr){6-8}          \cmidrule(lr){9-11}         \cmidrule(lr){12-14}        \cmidrule(lr){15-17}
                                &                       &                           &               &                           
                                                                            & R@1   & R@5   & R@10      & R@1  & R@5  & R@10        & R@1  & R@5  & R@10        & R@1  & R@5  & R@10    \\
    \midrule
        ViLT                    & -                     & 9.74                      & 573           & -                         
                                                                            & 83.5  & 96.7  & 98.6      & 64.4  & 88.7  & 93.8      & 61.5  & 86.3  & 92.7      & 42.7  & 72.9  & 83.1  \\
        UNITER                  & -                     & 0.20                      & 31            & -                         
                                                                            & 87.3  & 98.0  & 99.2      & 75.6  & 94.1  & 96.8      & 65.7  & 88.6  & 93.8      & 52.9  & 79.9  & 88.0  \\
        VILLA                   & -                     & $\sim$0.60                & $\sim$93      & -                         
                                                                            & 87.9  & 97.5  & 98.8      & 76.3  & 94.2  & 96.8      & -     & -     & -         & -     & -     & -     \\
        UNIMO                   & -                     & -                         & -             & -                         
                                                                            & 89.4  & 98.9  & 99.8      & 78.0  & 94.2  & 97.1      & -     & -     & -         & -     & -     & -     \\   
    \midrule
        METER                   & -                     & \textcolor{gray}{8.66}    &\textcolor{gray}{494} & \textcolor{gray}{90.0}    
                                                                            & \textcolor{gray}{89.6}    & \textcolor{gray}{98.3}    & \textcolor{gray}{99.4}    
                                                                                                        & \textcolor{gray}{77.0}    & \textcolor{gray}{94.5}    & \textcolor{gray}{97.5}    
                                                                                                                                    & -     & -     & -         & -     & -     & -     \\
        \hspace{2mm} $\mapsto$ EViT     & $\sim$25\%            & 4.68                      & 325           & 64.8                      
                                                                            & 60.5  & 86.6  & 92.6      & 44.9  & 77.4  & 86.6      & -     & -     & -         & -     & -     & -     \\
        \rowcolor{cyan!10}
        \hspace{2mm} $\mapsto$ ELIP     & $\sim$25\%            & 6.43                      & 420           & 70.4                      
                                                                            & 89.3  & 98.8  & 99.6      & 76.0  & 94.7  & 97.4      & -     & -     & -         & -     & -     & -     \\
    \midrule
        ALBEF                   & -                     & \textcolor{gray}{9.65}    &\textcolor{gray}{594}      & \textcolor{gray}{88.1}    
                                                                            & \textcolor{gray}{93.6}    & \textcolor{gray}{99.1}    & \textcolor{gray}{99.9}    
                                                                                                & \textcolor{gray}{81.0}    
                                                                                                        & \textcolor{gray}{96.0}    & \textcolor{gray}{97.8}    & \textcolor{gray}{72.2}    & \textcolor{gray}{91.8}    
                                                                                                        & \textcolor{gray}{96.1}    & \textcolor{gray}{55.9}    & \textcolor{gray}{81.4}    & \textcolor{gray}{88.8}    \\
        \hspace{2mm} $\mapsto$ EViT     & $\sim$30\%            & 3.21                      & 262            & 50.8                      & 87.7  & 97.8  & 98.6      & 75.4  & 93.1  & 96.7      & 65.7  & 88.4  & 94.0      & 49.7  & 77.1  & 85.8      \\
        \hspace{2mm} $\mapsto$ ToMe     & $\sim$30\%            & 6.66                      & 450            & 69.6                      & 92.1  & 98.7  & 99.6      & 78.1  & 94.6  & 97.6      & 68.8  & 90.1  & 94.9      & 51.9  & 79.1  & 87.1      \\
        \rowcolor{cyan!10}
        \hspace{2mm} $\mapsto$ ELIP     & $\sim$30\%            & 8.50                      & 518            & 69.6                      & 93.4  & 99.3  & 99.8      & 80.6  & 95.4  & 97.7      & 71.8  & 91.6  & 95.7      & 55.0  & 80.8  & 88.4      \\
    \midrule
        BLIP                    & -                     & \textcolor{gray}{11.03}   &\textcolor{gray}{1,102}                    & \textcolor{gray}{90.8}    & \textcolor{gray}{94.2}    & \textcolor{gray}{99.1}    & \textcolor{gray}{99.9}    
                                                                                                        & \textcolor{gray}{81.4}    
                                                                                                        & \textcolor{gray}{95.6}    & \textcolor{gray}{98.1}    & \textcolor{gray}{72.8}    & \textcolor{gray}{92.1}    
                                                                                                        & \textcolor{gray}{96.1}    & \textcolor{gray}{56.6}    & \textcolor{gray}{81.7}    & \textcolor{gray}{88.9}    \\
        \hspace{2mm} $\mapsto$ EViT     & $\sim$30\%            & 4.80                      & 536            & 60.8                      & 87.3  & 98.5  & 99.4      & 75.1  & 93.5  & 96.4      & 66.8  & 88.9  & 93.9      & 50.8  & 77.9  & 86.3      \\
        \hspace{2mm} $\mapsto$ ToMe     & $\sim$30\%            & 6.98                      & 740            & 72.0                      & 91.5  & 98.8  & 99.4      & 80.5  & 95.6  & 97.9      & 71.5  & 91.6  & 95.9      & 55.3  & 81.2  & 88.7      \\
        \rowcolor{cyan!10}
        \hspace{2mm} $\mapsto$ ELIP     & $\sim$30\%            & 9.34                      & 960            & 74.7                      & 92.2  & 99.1  & 99.7      & 80.3  & 96.0  & 98.0      & 72.0  & 91.9  & 95.9      & 56.3  & 81.2  & 88.7      \\                      
    \bottomrule
    \end{tabular}
    \end{adjustbox}
    \label{tab:retrieval}
\end{table*}

\begin{table}[htbp]
    \centering
    \caption{Performance comparison on VQA and $\text{NLVR}^2$ datasets.} 
    \begin{tabular}{l|cc|cc}
    \toprule
    \multirow{3}{*}{Model}              & \multicolumn{2}{c|}{VQA}                              & \multicolumn{2}{c}{$\text{NLVR}^2$}                   \\  
                                        \cmidrule(lr){2-3}                                      \cmidrule(lr){4-5}  
                                        & test-dev                  & test-std                  & dev                       & test-P                    \\
    \midrule
    VisualBERT                          & 70.80                     & 71.00                     & 67.40                     & 67.00                     \\
    ViLT                                & 71.26                     & -                         & 75.24                     & 76.21                     \\
    LXMERT                              & 72.42                     & 72.54                     & 74.90                     & 74.50                     \\
    UNITER                              & 72.70                     & 72.91                     & 77.18                     & 77.85                     \\
    12-in-1                             & 73.15                     & -                         & -                         & 78.87                     \\
    \midrule
    ALBEF                               & \textcolor{gray}{74.57}   & \textcolor{gray}{74.79}   & \textcolor{gray}{-}       & \textcolor{gray}{-}       \\
    \rowcolor{cyan!10}
    \hspace{2mm} $\mapsto$ ELIP                 & 74.33                     & 74.48                     & -                         & -                         \\
    \midrule
    METER                               & \textcolor{gray}{74.72}   & \textcolor{gray}{74.71}   & \textcolor{gray}{78.69}   & \textcolor{gray}{79.66}   \\
    \rowcolor{cyan!10}
    \hspace{2mm} $\mapsto$ ELIP                 & 74.16                     & 74.31                     & 78.41                     & 79.36                     \\
    \bottomrule
    \end{tabular}
    \label{tab:vqa}
\end{table}

\begin{table}[htbp]
    \centering
    \caption{Results of both val and test sets on SNLI-VE.} 
    \begin{tabular}{l|cc}
    \toprule
    Model                               & val                       & test                      \\
    \midrule
    12-in-1~\citep{12in1}                & -                         & 76.95                     \\
    UNITER~\citep{uniter}                & 78.59                     & 78.28                     \\ 
    \midrule
    ALBEF~\citep{albef}                  & \textcolor{gray}{79.33}   & \textcolor{gray}{79.41}   \\
    \hspace{2mm} $\mapsto$ EViT                 & 78.54                     & 78.75                     \\
    \hspace{2mm} $\mapsto$ ToMe                 & 78.69                     & 78.76                     \\
    \rowcolor{cyan!10}
    \hspace{2mm} $\mapsto$ ELIP                 & 79.24                     & 79.38                     \\
    \midrule
    METER~\citep{meter}                  & \textcolor{gray}{79.94}   & \textcolor{gray}{79.41}   \\
    \hspace{2mm} $\mapsto$ EViT                 & 74.65                     & 73.89                     \\
    \rowcolor{cyan!10}
    \hspace{2mm} $\mapsto$ ELIP                 & 79.59                     & 79.10                     \\
    \bottomrule
    \end{tabular}
    \label{tab:ve}
\end{table} 

\begin{table}[htbp]
    \centering
    \caption{Following BLIP, we compare only with a single representative baseline that has been pre-trained on large-scale datasets while excluding these conventional comparatively weak approaches.} 
    \begin{tabular}{l|c}
    \toprule
    Model                       & CIDEr$\uparrow$           \\
    \midrule
    Enc-Dec~\citep{cc12m}        & 110.9                     \\
    \midrule
    BLIP~\citep{blip}            & \textcolor{gray}{121.7}   \\
    \hspace{2mm} $\mapsto$ EViT         & 117.5                     \\
    \hspace{2mm} $\mapsto$ ToMe         & 121.4                     \\
    \rowcolor{cyan!10}
    \hspace{2mm} $\mapsto$ ELIP         & 122.5                     \\
    \bottomrule
    \end{tabular}
    \label{tab:ve-caption}
\end{table}

\subsection{Cross-Modal Learning Model Evaluation}
\subsubsection{More Settings}
\noindent\textbf{Pre-Training}
We utilized four publicly available large-scale datasets for pre-training: MSCOCO Caption~\citep{coco}, Visual Genome~\citep{genome}, SBU~\citep{sbu}, and Conceptual Captions~\citep{cc}, which together provide ~4M images.
We applied our ELIP method to three popular language-image pre-trained models, \ie ALBEF~\citep{albef}, BLIP~\citep{blip}, and METER~\citep{meter}.
For each individual model, we trained it from scratch using four NVIDIA A5000 GPUs and kept most of the experimental settings untouched, except for reducing the batch size due to resource constraints.

One exception about METER is that the number of ViT layers is eleven.
We thus adapted the number of Block III layers to five and set the retaining ratio to 70\%. 
Furthermore, we found it beneficial to decrease the retaining ratio to 45\% for the last block.

\noindent\textbf{Downstream VL Tasks}
We validated the effectiveness of our proposed method on five downstream vision-language tasks in this work.
\begin{itemize}[leftmargin=*]
    \item \textbf{Image-Text Retrieval}
    consists of two sub-tasks: image-to-text retrieval (\textbf{TR}) and text-to-image retrieval (\textbf{IR}).
    We tested the model performance on Flickr30K~\citep{flickr} and MSCOCO datasets~\citep{coco} using the recall metric \textbf{R}@$n$, \ie truncated top-$n$ results is employed.
    \item \textbf{Visual Entailment (SNLI-VE)}~\citep{ve} predicts the relationship of an image-text pair with three classes: entailment, neutral, or contradictory.
    We followed previous literature~\citep{uniter, albef} to treat this task as a three-way classification problem.
    \item \textbf{Visual Question Answering (VQA).} 
    We used the VQA v2 dataset~\citep{VQA2} and adopted accuracy as the key metric.
    Due to the submission number limitation of the leaderboard website, we merely evaluated the baseline and our model and reported the final results once.
    \item \textbf{Natural Language for Visual Reasoning ($\text{NLVR}^2$)}  requires a model to predict whether a given text describes a pair of images.
    We reported the accuracy of dev and test-P splits.
    \item \textbf{Image Captioning.} 
    BLIP is endowed with text generation capability compared to the other two models. 
    We therefore only conducted experiments on this task with BLIP using the MSCOCO Karpathy split setting~\citep{blip}.
\end{itemize}

\subsubsection{Overall Results}
In Table~\ref{tab:retrieval}, \ref{tab:vqa}, \ref{tab:ve}, and \ref{tab:ve-caption}, we present the performance comparison of our approach with other state-of-the-art methods on five downstream language-image tasks, involving six datasets in total.
The reported TFLOP values (both forward and backward) are estimated with a batch size of 36, and the GPU memory usage is calculated based on four A5000 GPUs (only backbone with and without ELIP method). 
We excluded some experiments due to resource reasons, \eg VQA on the BLIP approach, or incompatibilities, such as METER and ToMe.
The main observations are as follows:
\begin{itemize}[leftmargin=1em]
    \item Previous strong language-image pre-training methods, such as UNITER~\citep{uniter} and VILLA~\citep{villa} often employ pre-extracted object features (Faster RCNN) for vision encoder.
    While these approaches can be computationally less expensive in terms of TFLOPs, the retrieval results are often inferior to the recent models with ViT encoders such as ALBEF~\citep{albef} and BLIP~\citep{blip}.
    \item The two baseline methods -- EViT and ToMe, though reduce the model complexity by a large margin, often trade drastic model performance over the downstream tasks.
    For example, when applying EViT to the METER model, there is a significant drop of 20 to 30 points in R@1 for both text and image retrieval.
    \item Unlike the two baseline methods and other compared approaches, our ELIP model achieves a superior efficiency-effectiveness trade-off. 
    Specifically, across all five downstream tasks, our model yields an average accuracy drop of less than 0.4\% for the three backbone models, evidently demonstrating its effectiveness and generalization ability.
\end{itemize}

\subsection{Contrastive-Only Model Evaluation}
\subsubsection{More Settings}
Our primary objective in this study is to assess the efficacy of our proposed data-efficient method. 
Consequently, we did not conduct an extensive parameter search and instead utilized a universal setting across different models.
Due to limitations in computational resources, most of our pre-training experiments were conducted using four NVIDIA A5000 GPUs. 
Specifically, for CLIP models, we employed 32 epochs, a learning rate of 1e-3, and a weight decay of 0.1. 
The batch size for ViT-S/16 and ViT-B/16 is 1,760 and 920, respectively (We did not utilize gradient accumulation).
For the downstream image classification task, we fine-tuned the pre-trained models on a single NVIDIA A100-40G GPU. 
Fine-tuning comprises 10 epochs with a learning rate of 1e-3 and a weight decay of 0.1.

\noindent \textbf{Pre-Training Datasets.}
For \textbf{CLIP} models, we utilized two versions of pre-training datasets to examine the data-size scaling law as well. 
We employed the OpenCLIP repository~\citep{openclip} to \emph{conduct pre-training for all models from scratch}, ensuring a fair comparison between our proposed method and the baselines. 
Specifically, the smaller dataset, denoted as \textbf{CC3M+}, comprises CC3M~\citep{cc}, SBU-Captions~\citep{sbu}, and MSCOCO~\citep{coco}, totaling 4.1 million image-text pairs. 
The larger dataset, denoted as \textbf{CC12M+}, includes CC12M~\citep{cc12m}, SBU-Captions~\citep{sbu}, and MSCOCO~\citep{coco}, with a total of 10.1 million pairs.

\noindent \textbf{Downstream Datasets.}
We conducted extensive downstream fine-tuning experiments across various datasets. 
Specifically, we utilized datasets such as ImageNet~\citep{imagenet}, CIFAR-10, CIFAR-100~\citep{cifar}, as well as out-of-distribution datasets including ImageNet V2~\citep{imagenetv2} and ImageNet-R~\citep{imagenet-r} to validate the downstream performance of \textbf{CLIP} pre-trained models. 
For all these datasets, we explored diverse experimental settings, encompassing zero-shot transfer learning from ImageNet, linear probing, and full fine-tuning. 

\subsubsection{Overall Results}
Since ToMe is incompatible with the CLIP architecture, we compare our method only against EViT and the original CLIP backbone. 
As shown in Table~\ref{tab:clip-1} and Table~\ref{tab:clip-2}, our ELIP model consistently achieves a better efficiency-effectiveness trade-off than both CLIP and EViT.

\begin{table*}[htbp]
    \centering
    \caption{Comparison of \emph{full fine-tuning} downstream image classification tasks (IN-ImageNet, C10-CIFAR10, C100-CIFAR100).
    Two pre-training dataset versions are separately reported: \textbf{CC3M+} and \textbf{CC12M+}.
    Metrics pertaining to efficiency are estimated for pre-training rather than downstream fine-tuning.
    RRT: Ratio of Removed Tokens; Mem: Memory (GB).} 
    \begin{adjustbox}{width=\textwidth}
    \begin{tabular}{l|ccc *{2}{|ccccc}}
    \toprule
    \multirow{2}{*}{CLIP Arch}  & \multicolumn{3}{c|}{\violet{Effiency-Related}}    & \multicolumn{5}{c|}{\textbf{CC3M+}}               & \multicolumn{5}{c}{\textbf{CC12M+}}                           \\
                                \cmidrule(lr){2-4}                                  \cmidrule(lr){5-9}                                 \cmidrule(lr){10-14}
                                & \violet{RRT$\uparrow$}    & \violet{TFLOPs$\downarrow$}   & \violet{Mem$\downarrow$} 
                                                                                    & C10   & C100      & IN    & IN-V2     & IN-R      & C10   & C100      & IN    & IN-V2     & IN-R                  \\
    \midrule
    ViT-S/16                    & -                         & \textcolor{gray}{31.34}   & \textcolor{gray}{92.8}
                                                            & \textcolor{gray}{95.31}   & \textcolor{gray}{79.50}   & \textcolor{gray}{65.99}   & \textcolor{gray}{52.93}   & \textcolor{gray}{28.05}    
                                                            & \textcolor{gray}{95.89}   & \textcolor{gray}{82.33}   & \textcolor{gray}{69.43}   & \textcolor{gray}{56.97}   & \textcolor{gray}{33.08}   \\
    \hspace{2mm} $\mapsto$ EViT & $\sim$30\%                & 12.91                     & 65.2                      
                                                            & 94.04                     & 76.21                     & 64.50                     & 51.32                     & 24.07 
                                                            & 95.26                     & 80.04                     & 68.03                     & 54.99                     & 30.60                     \\
    \rowcolor{cyan!10}
    \hspace{2mm} $\mapsto$ ELIP & $\sim$30\%                & 23.98                     & 75.2
                                                            & 94.66                     & 77.58                     & 65.97                     & 52.66                     & 27.70
                                                            & 95.75                     & 82.09                     & 69.47                     & 56.53                     & 32.38                     \\
    \midrule
    ViT-B/16                    & -                         & \textcolor{gray}{97.41}   & \textcolor{gray}{93.6}    
                                                            & \textcolor{gray}{95.13}   & \textcolor{gray}{79.82}   & \textcolor{gray}{71.27}   & \textcolor{gray}{57.68}   & \textcolor{gray}{32.82}    
                                                            & \textcolor{gray}{95.84}   & \textcolor{gray}{81.31}   & \textcolor{gray}{71.44}   & \textcolor{gray}{58.28}   & \textcolor{gray}{32.60}   \\
    \hspace{2mm} $\mapsto$ EViT & $\sim$30\%                & 32.62                     & 63.2                      
                                                            & 94.45                     & 77.90                     & 67.85                     & 54.27                     & 26.56 
                                                            & 94.89                     & 79.61                     & 70.24                     & 57.09                     & 29.98                     \\
    \rowcolor{cyan!10}
    \hspace{2mm} $\mapsto$ ELIP & $\sim$30\%                & 72.21                     & 75.2                      
                                                            & 95.02                     & 79.59                     & 70.96                     & 56.92                     & 31.71 
                                                            & 95.50                     & 81.79                     & 71.41                     & 58.61                     & 32.86\\                      
    \bottomrule
    \end{tabular}
    \end{adjustbox}
    \label{tab:clip-1}
\end{table*}

\begin{table*}[htbp]
    \centering
    \caption{Comparison of \emph{zero-shot} and \emph{linear probing} downstream image classification tasks (IN-ImageNet, C10-CIFAR10, C100-CIFAR100).
    Two pre-training dataset versions are separately reported: \textbf{CC3M+} and \textbf{CC12M+}.
    } 
    \begin{adjustbox}{width=\textwidth}
    \begin{tabular}{l*{2}{|cc|ccccc}}
    \toprule
    \multirow{3}{*}{CLIP Arch}  & \multicolumn{7}{c|}{\textbf{CC3M+}}                                                       & \multicolumn{7}{c}{\textbf{CC12M+}}                                       \\
                                \cmidrule(lr){2-8}                                                                          \cmidrule(lr){9-15}
                                & \multicolumn{2}{c|}{IN Zero-Shot} & \multicolumn{5}{c|}{Linear Probing}                   & \multicolumn{2}{c|}{IN Zero-Shot} & \multicolumn{5}{c}{Linear Probing}   \\                  
                                \cmidrule(lr){2-3}                  \cmidrule(lr){4-8}                                      \cmidrule(lr){9-10}                 \cmidrule(lr){11-15}   
                                & top-1     & top-5                 & C10   & C100      & IN    & IN-V2     & IN-R          & top-1     & top-5 & C10   & C100      & IN    & IN-V2     & IN-R                              \\
    \midrule
    ViT-S/16                    & \textcolor{gray}{18.70}   & \textcolor{gray}{37.71}   & \textcolor{gray}{95.42}   & \textcolor{gray}{79.08}   & \textcolor{gray}{65.92}   & \textcolor{gray}{52.72}   & \textcolor{gray}{27.47}    
                                & \textcolor{gray}{34.78}   & \textcolor{gray}{62.42}   & \textcolor{gray}{96.10}   & \textcolor{gray}{82.64}   & \textcolor{gray}{69.37}   & \textcolor{gray}{56.36}   & \textcolor{gray}{32.92}\\
    \hspace{2mm} $\mapsto$ EViT & 15.12     & 31.86                 & 94.30 & 76.53     & 64.70 & 51.05     & 25.07         & 28.41 & 54.77     & 95.10 & 79.77     & 67.92 & 54.75     & 30.67         \\
    \rowcolor{cyan!10}
    \hspace{2mm} $\mapsto$ ELIP & 17.69     & 36.28                 & 94.53 & 77.94     & 64.89 & 51.79     & 26.86         & 33.78 & 61.70     & 95.83 & 82.06     & 69.15 & 56.05     & 32.27         \\
    \midrule
    ViT-B/16                    & \textcolor{gray}{20.65}   & \textcolor{gray}{40.82}   & \textcolor{gray}{95.32}   & \textcolor{gray}{79.96}   & \textcolor{gray}{68.90}   & \textcolor{gray}{55.59}   & \textcolor{gray}{28.44}
                                & \textcolor{gray}{35.78}   & \textcolor{gray}{63.47}   & \textcolor{gray}{95.76}   & \textcolor{gray}{81.38}   & \textcolor{gray}{70.07}   & \textcolor{gray}{57.16}   & \textcolor{gray}{30.56}\\
    \hspace{2mm} $\mapsto$ EViT & 17.48     & 35.99                 & 94.77 & 77.29     & 67.55 & 54.10     & 26.40         & 25.98 & 56.39     & 94.79 & 79.43     & 68.82 & 55.57     & 27.92\\
    \rowcolor{cyan!10}
    \hspace{2mm} $\mapsto$ ELIP & 20.63     & 40.48                 & 95.02 & 79.32     & 68.81 & 55.55     & 27.81         & 34.37 & 62.57     & 95.45 & 81.87     & 71.31 & 58.42     & 32.61\\                      
    \bottomrule
    \end{tabular}
    \end{adjustbox}
    \label{tab:clip-2}
\end{table*}

\begin{table}[t!]
    \centering
    \caption{The effect of three text pruning approaches on retrieval results of the Flickr30K dataset.} 
    \begin{tabular}{c|c|cc|cc}
    \toprule
    \multicolumn{2}{c|}{\multirow{2}{*}{Model}} & \multicolumn{2}{c|}{TR}   & \multicolumn{2}{c}{IR}    \\
                                                \cmidrule(lr){3-4}          \cmidrule(lr){5-6}
    \multicolumn{2}{c|}{}                       & R@1       & R@5           & R@1       & R@5           \\ 
    \midrule
    \multirow{4}{*}{METER}  & No Pruning        & \textcolor{gray}{84.2}    & \textcolor{gray}{97.7}
                                                & \textcolor{gray}{69.5}    & \textcolor{gray}{92.3}    \\
                            & Random            & 73.8      & 93.6          & 57.9      & 88.1          \\
                            & Post              & 76.1      & 94.6          & 61.6      & 89.0          \\
                            & Learned           & 76.9      & 94.8          & 60.9      & 89.2          \\
    \midrule
    \multirow{4}{*}{BLIP}   & No Pruning        & \textcolor{gray}{91.5}    & \textcolor{gray}{98.8}
                                                & \textcolor{gray}{77.9}    & \textcolor{gray}{94.6}    \\
                            & Random            & 89.9      & 98.2          & 75.8      & 94.1          \\
                            & Post              & 91.6      & 98.9          & 78.2      & 94.5          \\
                            & Learned           & 91.1      & 98.8          & 77.7      & 94.5          \\
    \bottomrule
    \end{tabular}
    \label{tab:text}
\end{table}

\begin{figure}[t!]
  \centering
  \includegraphics[width=1.0\linewidth]{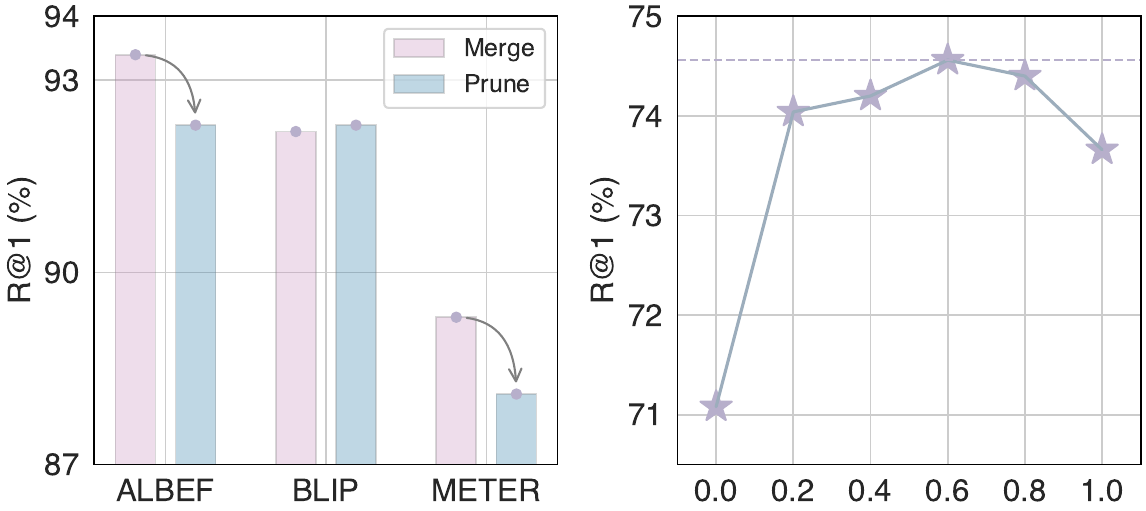}
  \caption{Component effect on the text retrieval performance over the Flickr30K dataset.
  Left: Performance comparison of pruning-only and pruning-then-merging approaches.
  Right: Performance change with respect to the feature combination coefficient parameter $\lambda$ in Eqn.~\ref{eqn:lambda}.}\label{fig:merge}
\end{figure}

\begin{table*}[htbp]
    \centering
    \caption{Ablation study of different pruning strategies on the Flickr30K dataset.
    OPR: Overall Pruning Ratio.} 
    \begin{adjustbox}{width=0.75\textwidth}
    \begin{tabular}{c|l|cc|cc}
    \toprule
    OPR$\downarrow$ & Retaining Strategy                                                            & TR@1  & TR@5  & IR@1  & IR@5  \\
    \midrule
    20\%            & 2(100\%) $\rightarrow$ 2(90\%) $\rightarrow$ 6(85\%) $\rightarrow$ 2(40\%)    & 88.1  & 98.2  & 75.2  & 93.3  \\
    \midrule
    \multirow{4}{*}{30\%}
                    & 6(100\%) $\rightarrow$ 6(40\%)                                                & 88.1  & 98.2  & 74.0  & 93.0  \\
                    & 2(100\%) $\rightarrow$ 2(90\%) $\rightarrow$ 8(55\%)                          & 88.1  & 98.4  & 73.9  & 92.8  \\
                    & 4(100\%) $\rightarrow$ 4(50\%) $\rightarrow$ 4(50\%)                          & 88.0  & 97.9  & 74.3  & 91.4  \\
                    & \textcolor{magenta}{2(100\%) $\rightarrow$ 2(90\%) $\rightarrow$ 6(65\%) $\rightarrow$ 2(40\%)} 
                    & \textcolor{magenta}{87.2}    & \textcolor{magenta}{98.4}    & \textcolor{magenta}{74.4}        & \textcolor{magenta}{92.9} \\
    \midrule
    40\%            & 2(100\%) $\rightarrow$ 2(90\%) $\rightarrow$ 6(40\%) $\rightarrow$ 2(40\%)    & 86.7  & 98.1  & 73.3  & 92.8  \\
    \bottomrule
    \end{tabular}
    \end{adjustbox}
    \label{tab:ratios}
\end{table*}


\subsection{Ablation Study}

\begin{table*}[htbp]
    \centering
    \caption{Performance of scaling ELIP pre-training to larger batch sizes.
    We also show the relative improvement of $\text{ELIP}_{\textcolor{blue}{+}}$ (larger batch size under the same budget with Base models) over the original Base models on the retrieval results.} 
    \begin{adjustbox}{width=0.8\textwidth}
    \begin{tabular}{c|l|c|ccc|ccc}
    \toprule
    \multicolumn{2}{c|}{\multirow{2}{*}{Model}}                     & \multirow{2}{*}{batch size$\uparrow$} & \multicolumn{3}{c|}{TR}   & \multicolumn{3}{c}{IR}    \\
                                                                                                            \cmidrule(lr){4-6}          \cmidrule(lr){7-9}
    \multicolumn{2}{c|}{}                                           &                                       & R@1   & R@5   & R@10      & R@1  & R@5  & R@10        \\ 
    \midrule
    \multirow{3}{*}{METER}  & Base                                  & 36$\times$4                           & \textcolor{gray}{89.6}    & \textcolor{gray}{98.3}    
                                                                                                            & \textcolor{gray}{99.4}    & \textcolor{gray}{77.0}    
                                                                                                            & \textcolor{gray}{94.5}    & \textcolor{gray}{97.5}    \\
                            & ELIP                                  & 36$\times$4                           & 89.3  & 98.8  & 99.6      & 76.0  & 94.7  & 97.4      \\
                            & $\text{ELIP}_{\textcolor{blue}{+}}$   & 54$\times$4                           & 88.7  & 98.4$_{\textcolor{blue}{+.1}}$  & 99.4      
                                                                                                            & 75.8  & 94.2  & 97.2      \\
    \midrule
    \multirow{3}{*}{ALBEF}  & Base                                  & 40$\times$4                           & \textcolor{gray}{93.6}    & \textcolor{gray}{99.1}    & \textcolor{gray}{99.9}    
                                                                                                            & \textcolor{gray}{81.0}    & \textcolor{gray}{96.0}    & \textcolor{gray}{97.8}    \\
                            & ELIP                                  & 40$\times$4                           & 93.4  & 99.3  & 99.8      & 80.6  & 95.4  & 97.7      \\
                            & $\text{ELIP}_{\textcolor{blue}{+}}$   & 58$\times$4                           & 93.7$_{\textcolor{blue}{+.1}}$    & 99.3$_{\textcolor{blue}{+.2}}$	
                                                                                                            & 100.0$_{\textcolor{blue}{+.1}}$	& 81.1$_{\textcolor{blue}{+.1}}$	
                                                                                                            & 95.6                              & 98.0$_{\textcolor{blue}{+.2}}$ \\   
    \midrule
    \multirow{3}{*}{BLIP}   & Base                                  & 42$\times$4                           & \textcolor{gray}{94.2}    & \textcolor{gray}{99.1}    & \textcolor{gray}{99.9}    
                                                                                                            & \textcolor{gray}{81.4}    & \textcolor{gray}{95.6}    & \textcolor{gray}{98.1}    \\
                            & ELIP                                  & 42$\times$4                           & 92.2  & 99.1  & 99.7      & 80.3  & 96.0  & 98.0      \\
                            & $\text{ELIP}_{\textcolor{blue}{+}}$   & 56$\times$4                           & 92.7  & 99.2$_{\textcolor{blue}{+.1}}$  & 99.7      
                                                                                                            & 80.7  & 95.4  & 98.0      \\
    \bottomrule
    \end{tabular}
    \end{adjustbox}
    \label{tab:scale}
\end{table*}

\subsubsection{Text Token Removal}
In typical language-image pre-training datasets, such as Conceptual Captions~\citep{cc}, the text is often accompanied by a short context, consisting of approximately 20 words per sentence. 
Moreover, previous studies have shown that language tokens are typically less redundant and have a higher information density in their representation~\citep{maskAE, flip}. 
As a result, in our method, we did not introduce text token pruning and only performed ablation experiments to investigate its influence on model performance.
Specifically, we preserved the first half of all Transformer layers and pruned 40\% of text tokens in the second half to achieve a balance between efficiency and effectiveness.

We designed three pruning strategies for this experiment: 
\emph{Random} -- The tokens are randomly pruned; 
\emph{Post} -- We prioritized pruning the post tokens; 
and \emph{Learned} -- We used the [CLS] token features from the vision encoder to guide the pruning of text tokens.
We run these models for \textbf{three} pre-training epochs and report the results in Table~\ref{tab:text}.
One can observe that: 
I) Among the three approaches, random pruning tends to perform unsatisfactorily due to the unexpected removal of crucial text tokens, leading to inferior performance.
II) Pertaining to text token pruning, the BLIP model is less affected than the METER model. 
For instance, with the post-pruning approach, the BLIP model's performance even slightly surpasses that of the non-pruned model.

\begin{table*}[htbp]
    \centering
    \caption{Model results after combining with several parameter-efficient transfer learning techniques.} 
    \begin{adjustbox}{width=0.9\textwidth}
    \begin{tabular}{c|c|c|cccc|cccc|c}
    \toprule
    \multicolumn{3}{c|}{\multirow{2}{*}{Model}}         & \multicolumn{4}{c|}{Flickr30K}    & \multicolumn{4}{c|}{MSCOCO}   & \multirow{2}{*}{Mem}                  \\
                                                        \cmidrule(lr){4-7}                  \cmidrule(lr){8-11}
    \multicolumn{3}{c|}{}                               & TR@1  & TR@5  & IR@1  & IR@5      & TR@1  & TR@5  & IR@1  & IR@5  &                                       \\ 
    \midrule
    \multirow{4}{*}{ALBEF}  & \multicolumn{2}{c|}{Base} & \textcolor{gray}{93.6}    & \textcolor{gray}{99.1}    & \textcolor{gray}{81.0}    & \textcolor{gray}{96.0}    
                                                        & \textcolor{gray}{72.2}    & \textcolor{gray}{91.8}    & \textcolor{gray}{55.9}    & \textcolor{gray}{81.4}    
                                                                                                                                            & \textcolor{gray}{172.8} \\
                            \cmidrule(lr){2-12}    
                            &\multirow{3}{*}{ELIP}
                            & AMP                       & 93.1  & 99.5  & 80.5  & 95.4      & 72.0  & 91.5  & 55.3  & 80.8  & 147.6                                 \\ 
                            && Adapter                  & 92.3  & 99.3  & 80.2  & 95.2      & 71.8  & 91.0  & 55.2  & 80.9  & 143.6                                 \\
                            && LoRA                     & 92.8  & 99.0  & 79.8  & 95.2      & 71.8  & 91.2  & 55.4  & 81.1  & 142.4                                 \\
    \midrule
    \multirow{4}{*}{BLIP}   & \multicolumn{2}{c|}{Base} & \textcolor{gray}{94.2}    & \textcolor{gray}{99.1}    & \textcolor{gray}{81.4}    & \textcolor{gray}{95.6}    
                                                        & \textcolor{gray}{72.8}    & \textcolor{gray}{92.1}    & \textcolor{gray}{56.6}    & \textcolor{gray}{81.7}    
                                                                                                                                            & \textcolor{gray}{168.4} \\
                            \cmidrule(lr){2-12} 
                            &\multirow{3}{*}{ELIP}
                            & AMP                       & 92.3  & 99.1  & 80.2  & 95.8      & 72.4  & 92.2  & 56.6  & 81.6  & 152.3                 \\ 
                            && Adapter                  & 91.1  & 99.1  & 80.3  & 95.8      & 72.2  & 91.7  & 55.9  & 81.1  & 152.3                 \\
                            && LoRA                     & 92.1  & 98.8  & 79.5  & 95.2      & 72.3  & 91.6  & 55.9  & 81.3  & 149.5                 \\                 
    \bottomrule
    \end{tabular}
    \end{adjustbox}
    \label{tab:peft}
\end{table*}


\subsubsection{Token Merging \textit{v.s.} Pruning}
One alternative way to deal with the inattentive tokens is to directly prune them.
Note that this pruning-only strategy leads to a slight efficiency improvement compared to the merging one.
To study its effectiveness in downstream fine-tuning, we removed the merging operation in Eqn.~\ref{eqn:merge} and observed the performance change of this model.
As shown in Fig.~\ref{fig:merge}, we can see that compared with the token merging, the pruning-only strategy usually results in inferior downstream performance.
This finding implies that these less attentive tokens still positively contribute to the final model performance, and removing them, especially from the input image space (as proposed in~\citep{flip}), may lead to sub-optimal downstream fine-tuning results.

\subsubsection{Effect of Coefficient Parameter $\lambda$}
We also experimented with different coefficient values in Eqn.~\ref{eqn:lambda}.
We conducted this experiment with three pre-training epochs and show the downstream performance change using different $\lambda$ in Fig.~\ref{fig:merge}.
The figure indicates that using the vision or text [CLS] token only for the supervision of token pruning leads to inferior outcomes.
On the other hand, the combination of these two, \ie when $\lambda=0.6$ consistently outperforms the other values tested.
This result supports the validity of leveraging multi-modal feature interaction for token pruning and merging in our proposed ELIP method.

\subsubsection{Ablation Study of Retaining Ratios}
We presented the results of several variants of our method, each employing a unique retaining strategy, in Table~\ref{tab:ratios}. 
Each model was pre-trained over three epochs, and their performance was subsequently evaluated using the Flickr30K dataset. 
It can be observed that a higher OPR generally results in improved model performance, as evidenced by the comparison between 20\% and 40\%. 
Moreover, we chose the row with blue color for both its effectiveness and alignment with existing hierarchical vision transformers~\citep{swin}.


\subsubsection{Downstream Pruning}
We performed pruning and merging on downstream fine-tuning retrieval tasks and reported the results in Table~\ref{tab:retrieval}.
It can be seen that pruning on downstream tasks often leads to further performance drops.
This result can be attributed to the complex interplay between the image and text modalities, which require a more nuanced understanding of contextual features to achieve high accuracy.  

\subsection{Pre-Training Scaling}
Our method uses fewer vision tokens during training compared to baseline models, allowing us to spare GPU memory and scale the model to larger batch sizes. 
To study this effect, we carefully increased the pre-training batch size while ensuring that the required GPU memory remained less than the original pre-training. 
Besides, we also estimated the latency of each pre-training epoch.
We performed this test on the Flickr30K dataset and illustrated the results in Table~\ref{tab:scale}.
Our observations for this result are three-fold:
\begin{itemize}[leftmargin=1em]
    \item Our ELIP method is able to maintain performance similar to the base model, while also accelerating the pre-training process and reducing the required GPU memory usage.
    \item The spared GPU memory enables us to scale the pre-training with larger batch sizes, \ie $\text{ELIP}_{\textcolor{blue}{+}}$ approach.
    For example, with METER, we increased the batch size from 36$\times$4 to 54$\times$4, resulting in a significant improvement in training efficiency, and a reduction in BLIP pre-training time by approximately 15\%.
    \item In terms of fine-tuning, our $\text{ELIP}_{\textcolor{blue}{+}}$ surpasses the ELIP by a large margin, and even slightly outperforms the base model in some cases.
    These results are rather promising as scaling model pre-training brings significant improvement in both downstream performance and efficiency.
\end{itemize}

\subsection{Results with PEFT Approaches}
We also evaluated the effectiveness of combining our pre-training methods with several PEFT approaches. 
As presented in Table~\ref{tab:peft}, our method achieves a more compatible combination with PEFT approaches, such as LoRA.
This combination leads to a reduction in consumed memory while maintaining comparable performance with that of the original backbones.


\subsection{Visualization of Token Pruning}
\begin{figure*}[htbp]
  \centering
  \includegraphics[width=0.9\linewidth]{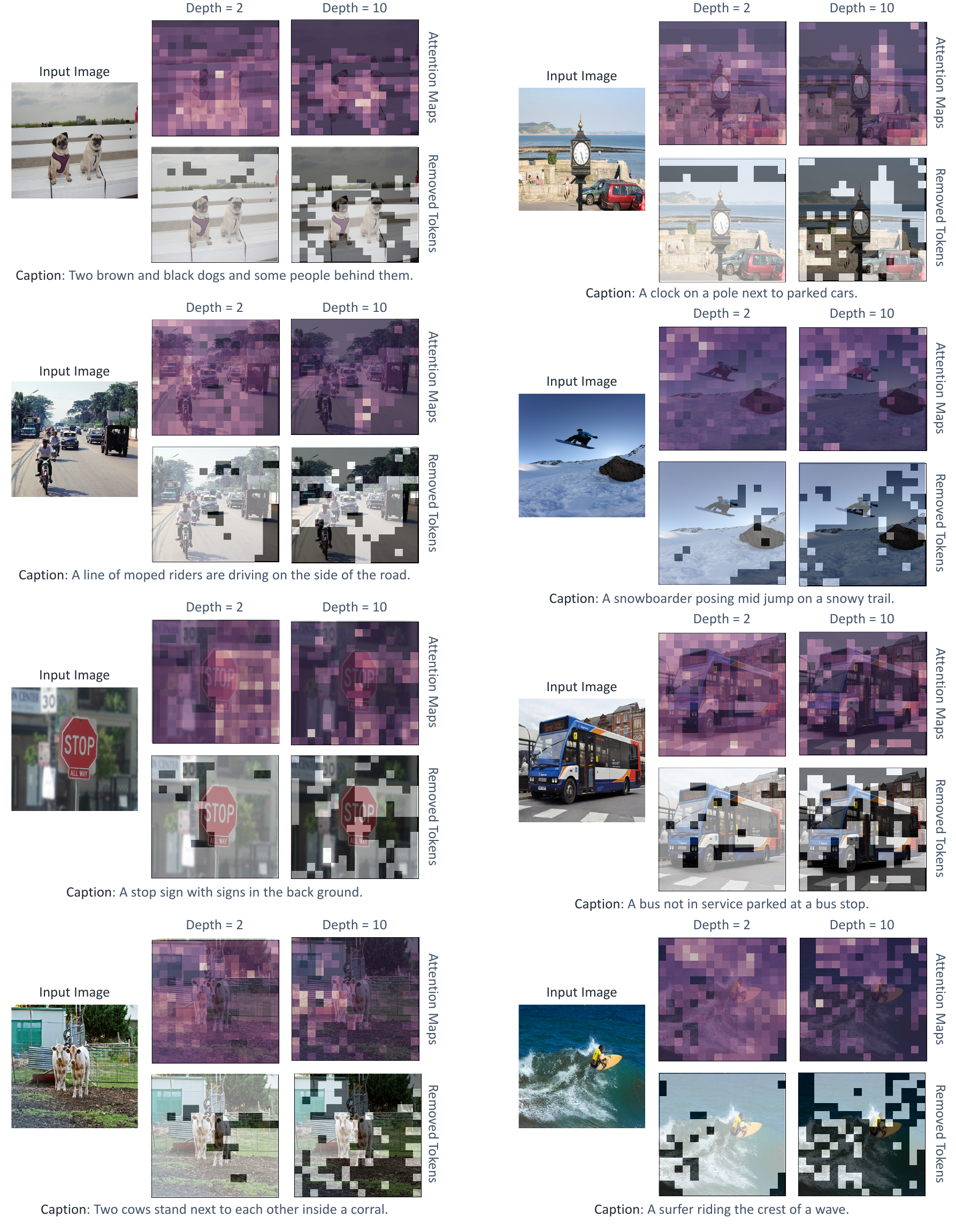}
  \caption{Visualization of pruning results with respect to two ViT depths: 2 and 10.
  Note that the effective vision tokens are gradually decreased by our method.
  We omit the merged tokens and show only the attention maps of the remaining ones for a clear illustration.
  }\label{fig:viz}
\end{figure*}

As illustrated in Sec.~\ref{sec:architecture}, our method consists of four blocks, wherein we perform pruning and merging in the last three blocks.
To quantitatively demonstrate the effectiveness of our pruning approach, we randomly selected two cases and presented them in Fig.~\ref{fig:viz}.
In particular, we mainly show the pruned attention map for two ViT layers: 2 and 10, and the effective vision tokens are gradually reduced with deeper ViT layers.
From this figure, we can observe that our method progressively removes less important vision tokens with deeper ViT layers.
For example, in the first case, the model gradually filters out the background information and puts more attention on the two \emph{dogs}.

\section{Conclusion and Future Work}\label{sec:conclusion}
In this paper, we propose a  token merging approach to achieve efficient discriminative language-image pre-training without introducing any additional trainable parameters.
We progressively prune and merge less influential vision tokens based on the language output using multiple stages.
Despite its simplicity, we show that our approach helps remove $\sim$30\% vision tokens whilst maintaining comparable performance with backbones over diverse downstream fine-tuning tasks.
Our method offers valuable insights for future research in discriminative language-image pre-training under limited computing resources, and may potentially benefit other multi-modal pre-training tasks such as video-language pre-training.

While our method demonstrates effectiveness in efficiency and scalability, one limitation is the lack of flexibility in the pruning ratio definition.
Therefore, an adaptive approach may be more helpful as different images often exhibit varying degrees of information sparsity.
In addition, our method can be seamlessly integrated with other efficient techniques, such as mixed-precision computation and gradient checkpointing, and thus build an even more efficient and lightweight language-image pre-training model.

{\small
\bibliographystyle{ieeenat_fullname}
\bibliography{VL-Pruning}
}

\end{document}